\documentclass[11pt,a4paper]{article}
\usepackage[hyperref]{acl2021}
\usepackage{times}
\usepackage{latexsym}
\usepackage{graphics}
\usepackage{graphicx}
\usepackage{color}
\usepackage{colortbl}
\usepackage{amssymb}
\usepackage{amsmath}
\usepackage{bbm}
\usepackage{booktabs}

\usepackage{microtype}
\usepackage{multirow}
\usepackage{footmisc}
\usepackage{soul}
\usepackage{arydshln}

\aclfinalcopy % Uncomment this line for the final submission
\def\aclpaperid{1280} %  Enter the acl Paper ID here

\title{{C}on{SERT}: {A} Contrastive Framework for Self-Supervised Sentence Representation Transfer} % 初步，待定

\author{
Yuanmeng Yan$^{1}$\thanks{\ \ Work done during internship at Meituan Inc. The first two authors contribute equally. Weiran Xu is the corresponding author.} ,
Rumei Li$^{2*}$,
Sirui Wang$^2$,
Fuzheng Zhang$^2$,
Wei Wu$^2$,
Weiran Xu$^1$\\
$^1$Beijing University of Posts and Telecommunications, Beijing, China\\
$^2$Meituan Inc., Beijing, China \\
\texttt{\{yanyuanmeng,xuweiran\}@bupt.edu.cn}\\
\texttt{\{lirumei,wangsirui,zhangfuzheng,wuwei30\}@meituan.com}
}

\date{}

\begin{document}
\maketitle
\begin{abstract}
Learning high-quality sentence representations benefits a wide range of natural language processing tasks. Though BERT-based pre-trained language models achieve high performance on many downstream tasks, the native derived sentence representations are proved to be collapsed and thus produce a poor performance on the semantic textual similarity (STS) tasks. In this paper, we present {C}on{SERT}, a \textbf{Con}trastive Framework for Self-Supervised \textbf{SE}ntence \textbf{R}epresentation \textbf{T}ransfer, that adopts contrastive learning to fine-tune BERT in an unsupervised and effective way. By making use of unlabeled texts, ConSERT solves the collapse issue of BERT-derived sentence representations and make them more applicable for downstream tasks. Experiments on STS datasets demonstrate that ConSERT achieves an 8\% relative improvement over the previous state-of-the-art, even comparable to the supervised SBERT-NLI. And when further incorporating NLI supervision, we achieve new state-of-the-art performance on STS tasks. Moreover, ConSERT obtains comparable results with only 1000 samples available, showing its robustness in data scarcity scenarios.
\end{abstract}

\section{Introduction}

% 介绍任务（句子表示学习）的背景、意义
Sentence representation learning plays a vital role in natural language processing tasks \cite{kiros2015skip, hill2016learning, conneau2017supervised, cer2018universal}.
Good sentence representations benefit a wide range of downstream tasks, especially for computationally expensive ones, including large-scale semantic similarity comparison and information retrieval.

\begin{figure}
    \centering
    \resizebox{.48\textwidth}{!}{
    \includegraphics{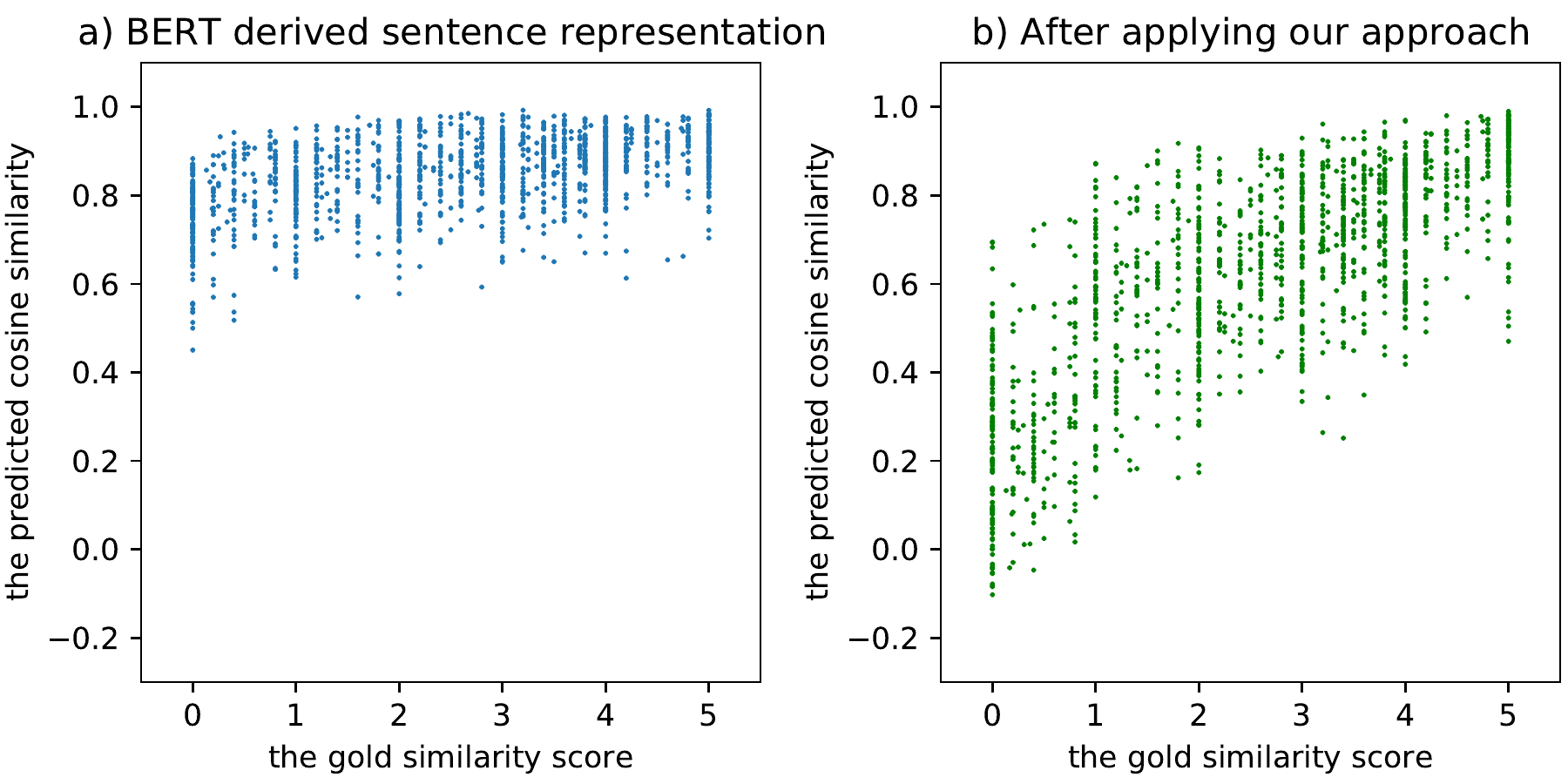}}
    \caption{The correlation diagram between the gold similarity score (x-axis) and the model predicted cosine similarity score (y-axis) on the STS benchmark dataset.
    % Following \citet{li-etal-2020-sentence}, the sentence representations are obtained by averaging token embeddings at the last two layers.
    }
    \label{fig:intro}
    \vspace{-0.5cm}
\end{figure}

Recently, BERT-based pre-trained language models have achieved high performance on many downstream tasks with additional supervision. However, the native sentence representations derived from BERT\footnote{Typically, we take the output of the [CLS] token or average token embeddings at the last few layers as the sentence representations.} are proved to be of low-quality \cite{reimers2019sentencebert, li-etal-2020-sentence}. As shown in Figure \ref{fig:intro}a, when directly adopt BERT-based sentence representations to semantic textual similarity (STS) tasks, almost all pairs of sentences achieved a similarity score between 0.6 to 1.0
%regardless of their gold similarity score (x axis)
, even if some pairs are regarded as completely unrelated by the human annotators. 
In other words, the BERT-derived native sentence representations are somehow collapsed \cite{chen2020exploring}, which means almost all sentences are mapped into a small area and therefore produce high similarity.

Such phenomenon is also observed in several previous works \cite{gao2019representation, wang2019improving, li-etal-2020-sentence}. They find the word representation space of BERT is anisotropic, the high-frequency words are clustered and close to the origin, while low-frequency words disperse sparsely. When averaging token embeddings, those high-frequency words dominate the sentence representations, inducing biases against their real semantics \footnote{We also empirically prove this hypothesis, please refer to Section \ref{sec:analysis_of_bert_embedding_space} for more details.}. 
As a result, it is inappropriate to directly apply BERT's native sentence representations for semantic matching or text retrieval. Traditional methods usually fine-tune BERT with additional supervision. However, human annotation is costly and often unavailable in real-world scenarios.

To alleviate the collapse issue of BERT as well as reduce the requirement for labeled data, we propose a novel sentence-level training objective based on contrastive learning \cite{he2020momentum, chen2020simple, chen2020big}. By encouraging two augmented views from the same sentence to be closer while keeping views from other sentences away, we reshape the BERT-derived sentence representation space and successfully solve the collapse issue (shown in Figure \ref{fig:intro}b). Moreover, we propose multiple data augmentation strategies for contrastive learning, including adversarial attack \cite{goodfellow2014explaining, kurakin2016adversarial}, token shuffling, cutoff \cite{shen2020simple} and dropout \cite{hinton2012improving}, that effectively transfer the sentence representations to downstream tasks. We name our approach ConSERT, a \textbf{Con}trastive Framework for \textbf{SE}ntence \textbf{R}epresentation \textbf{T}ransfer.

ConSERT has several advantages over previous approaches. Firstly, it introduces no extra structure or specialized implementation during inference. The parameter size of ConSERT keeps the same as BERT, making it easy to use. Secondly, compared with pre-training approaches, ConSERT is more efficient. With only 1,000 unlabeled texts drawn from the target distribution (which is easy to collect in real-world applications), we achieve 35\% relative performance gain over BERT, and the training stage takes only a few minutes (1-2k steps) on a single V100 GPU. Finally, it includes several effective and convenient data augmentation methods with minimal semantic impact. Their effects are validated and analyzed in the ablation studies.

Our contributions can be summarized as follows: 1) We propose a simple but effective sentence-level training objective based on contrastive learning. It mitigates the collapse of BERT-derived representations and transfers them to downstream tasks. 2) We explore various effective text augmentation strategies to generate views for contrastive learning and analyze their effects on unsupervised sentence representation transfer. 3) With only fine-tuning on unsupervised target datasets, our approach achieves significant improvement on STS tasks. When further incorporating with NLI supervision, our approach achieves new state-of-the-art performance. We also show the robustness of our approach in data scarcity scenarios and intuitive analysis of the transferred representations.\footnote{Our code is available at \url{https://github.com/yym6472/ConSERT}.}

\section{Related Work}
\label{sec:related_works}

\subsection{Sentence Representation Learning}

\textbf{Supervised Approaches} Several works use supervised datasets for sentence representation learning. \citet{conneau2017supervised} finds the supervised Natural Language Inference (NLI) task is useful to train good sentence representations. They use a BiLSTM-based encoder and train it on two NLI datasets, Stanford NLI (SNLI) \cite{bowman2015large} and Multi-Genre NLI (MNLI) \cite{williams2018broad}. Universal Sentence Encoder \cite{cer2018universal} adopts a Transformer-based architecture and uses the SNLI dataset to augment the unsupervised training. SBERT \cite{reimers2019sentencebert} proposes a siamese architecture with a shared BERT encoder and is also trained on SNLI and MNLI datasets.

\textbf{Self-supervised Objectives for Pre-training} BERT \cite{devlin2019bert} proposes a bi-directional Transformer encoder for language model pre-training. It includes a sentence-level training objective, namely next sentence prediction (NSP), which predicts whether two sentences are adjacent or not. However, NSP is proved to be weak and has little contribution to the final performance \cite{liu2019roberta}. After that, various self-supervised objectives are proposed for pre-training BERT-like sentence encoders. Cross-Thought \cite{wang2020cross} and CMLM \cite{yang2020universal} are two similar objectives that recover masked tokens in one sentence conditioned on the representations of its contextual sentences. SLM \cite{lee2020slm} proposes an objective that reconstructs the correct sentence ordering given the shuffled sentences as the input. However, all these objectives need document-level corpus and are thus not applicable to downstream tasks with only short texts.

\textbf{Unsupervised Approaches} BERT-flow \cite{li-etal-2020-sentence} proposes a flow-based approach that maps BERT embeddings to a standard Gaussian latent space, where embeddings are more suitable for comparison. However, this approach introduces extra model structures and need specialized implementation, which may limit its application.

\subsection{Contrastive Learning}

\textbf{Contrastive Learning for Visual Representation Learning} Recently, contrastive learning has become a very popular technique in unsupervised visual representation learning with solid performance \cite{chen2020simple, he2020momentum, chen2020big}. They believe that good representation should be able to identify the same object while distinguishing itself from other objects. Based on this intuition, they apply image transformations (e.g. cropping, rotation, cutout, etc.) to randomly generate two augmented versions for each image and make them close in the representation space. Such approaches can be regarded as the invariance modeling to the input samples. \citet{chen2020simple} proposes SimCLR, a simple framework for contrastive learning. They use the normalized temperature-scaled cross-entropy loss (NT-Xent) as the training loss, which is also called InfoNCE in the previous literature \cite{hjelm2018learning}.

% IS-BERT, DeCLUTR, CERT, CAPT, CLEAR, CT
\textbf{Contrastive Learning for Textual Representation Learning} Recently, contrastive learning has been widely applied in NLP tasks. Many works use it for language model pre-training. IS-BERT \cite{zhang2020unsupervised} proposes to add 1-D convolutional neural network (CNN) layers on top of BERT and train the CNNs by maximizing the mutual information (MI) between the global sentence embedding and its corresponding local contexts embeddings. CERT \cite{fang2020cert} adopts a similar structure as MoCo \cite{he2020momentum} and uses back-translation for data augmentation. However, the momentum encoder needs extra memory and back-translation may produce false positives. BERT-CT \cite{carlsson2021semantic} uses two individual encoders for contrastive learning, which also needs extra memory. Besides, they only sample 7 negatives, resulting in low training efficiency. DeCLUTR \cite{Giorgi2020DeCLUTRDC} adopts the architecture of SimCLR and jointly trains the model with contrastive objective and masked language model objective. However, they only use spans for contrastive learning, which is fragmented in semantics. CLEAR \cite{wu2020clear} uses the same architecture and objectives as DeCLUTR. Both of them are used to pre-train the language model, which needs a large corpus and takes a lot of resources.

\section{Approach}
\label{sec:approach}

In this section, we present ConSERT for sentence representation transfer. Given a BERT-like pre-trained language model $\mathbf{M}$ and an unsupervised dataset $\mathcal{D}$ drawn from the target distribution, we aim at fine-tuning $\mathbf{M}$ on $\mathcal{D}$ to make the sentence representation more task-relevant and applicable to downstream tasks. We first present the general framework of our approach, then we introduce several data augmentation strategies for contrastive learning. Finally, we talk about three ways to further incorporate supervision signals.

\subsection{General Framework}

\begin{figure}[t]
    \centering
    \resizebox{.48\textwidth}{!}{
    \includegraphics{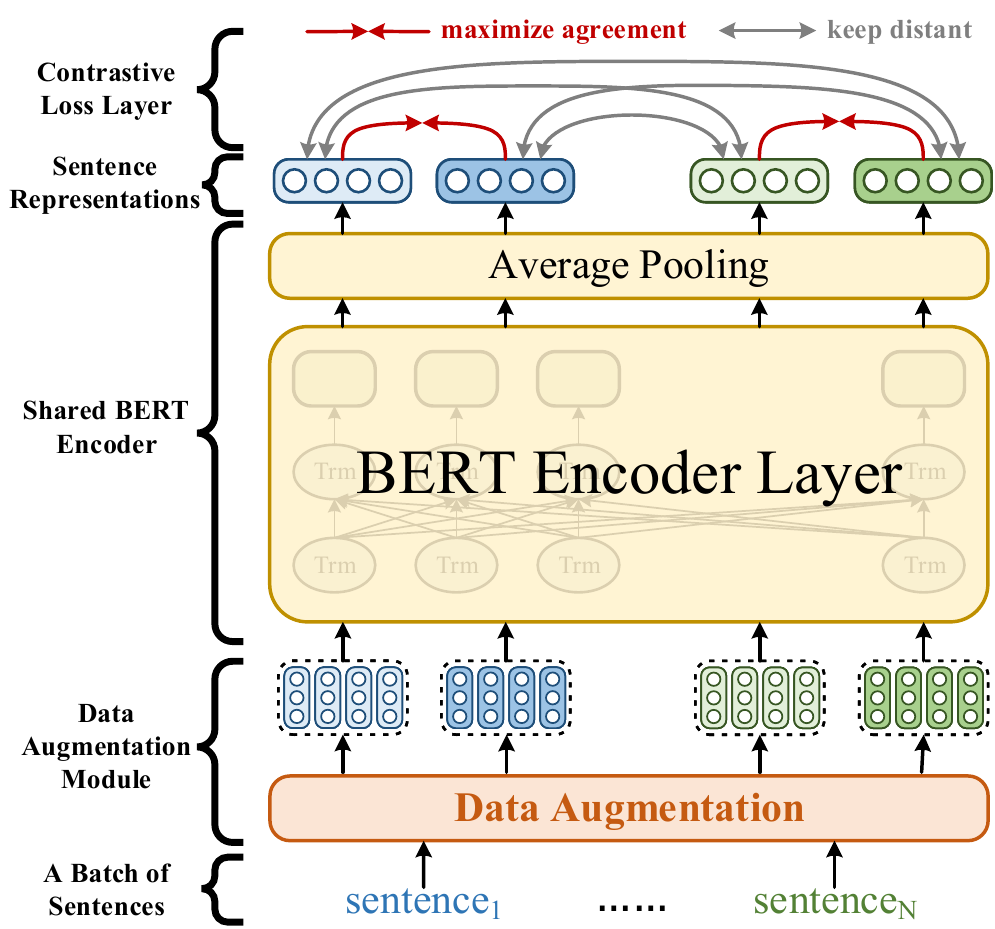}}
    \caption{The general framework of our proposed approach.}
    \label{fig:framework}
    \vspace{-0.5cm}
\end{figure}

\begin{figure*}[t]
    \centering
    \resizebox{.9\textwidth}{!}{
    \includegraphics{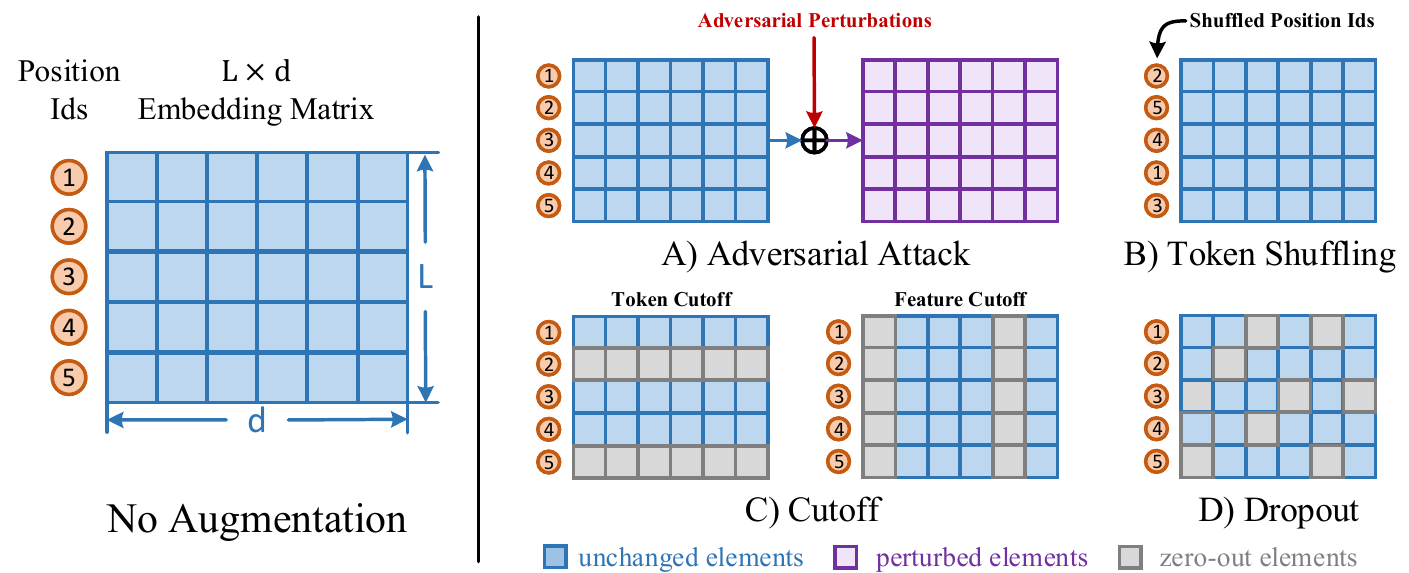}}
    \caption{The four data augmentation strategies used in our experiments.}
    \label{fig:data_aug}
    \vspace{-0.3cm}
\end{figure*}

Our approach is mainly inspired by SimCLR \cite{chen2020simple}. As shown in Figure \ref{fig:framework}, there are three major components in our framework:
\begin{itemize}
    \item A data augmentation module that generates different views for input samples at the token embedding layer.
    \item A shared BERT encoder that computes sentence representations for each input text. During training, we use the average pooling of the token embeddings at the last layer to obtain sentence representations.
    \item A contrastive loss layer on top of the BERT encoder. It maximizes the agreement between one representation and its corresponding version that is augmented from the same sentence while keeping it distant from other sentence representations in the same batch.
\end{itemize}

For each input text $x$, we first pass it to the data augmentation module, in which two transformations $T_1$ and $T_2$ are applied to generate two versions of token embeddings: $\boldsymbol{e}_i = T_1(x),  \boldsymbol{e}_j = T_2(x)$, where $\boldsymbol{e}_i, \boldsymbol{e}_j \in \mathbb{R}^{L \times d}$, $L$ is the sequence length and $d$ is the hidden dimension.
After that, both $\boldsymbol{e}_i$ and $\boldsymbol{e}_j$ will be encoded by multi-layer transformer blocks in BERT and produce the sentence representations $r_i$ and $r_j$ through average pooling.

Following \citet{chen2020simple}, we adopt the normalized temperature-scaled cross-entropy loss (NT-Xent) as the contrastive objective. During each training step, we randomly sample $N$ texts from $\mathcal{D}$ to construct a mini-batch, resulting in $2N$ representations after augmentation. Each data point is trained to find out its counterpart among $2(N-1)$ in-batch negative samples:
\begin{align}
    \setlength{\abovedisplayskip}{0.10cm}
    \setlength{\belowdisplayskip}{0.10cm}
    \mathcal{L}_{i, j} = - \log \frac{\exp(\text{sim}(r_i, r_j)/\tau)}{\sum_{k=1}^{2N} \mathbbm{1}_{[k \neq i]}\exp(\text{sim}(r_i, r_k)/\tau)}
    \label{equation:cl_loss}
\end{align}
, where $\text{sim}(\cdot)$ indicates the cosine similarity function, $\tau$ controls the temperature and $\mathbbm{1}$ is the indicator. Finally, we average all $2N$ in-batch classification losses to obtain the final contrastive loss $\mathcal{L}_\text{con}$.

\subsection{Data Augmentation Strategies}

We explore four different data augmentation strategies to generate views for contrastive learning, including adversarial attack \cite{goodfellow2014explaining, kurakin2016adversarial}, token shuffling, cutoff \cite{shen2020simple} and dropout \cite{hinton2012improving}, as illustrated in Figure \ref{fig:data_aug}.

\textbf{Adversarial Attack} Adversarial training is generally used to improve the model's robustness. They generate adversarial samples by adding a worst-case perturbation to the input sample. We implement this strategy with Fast Gradient Value (FGV) \cite{rozsa2016adversarial}, which directly uses the gradient to compute the perturbation and thus is faster than two-step alternative methods. Note that this strategy is only applicable when jointly training with supervision since it relies on supervised loss to compute adversarial perturbations.

\textbf{Token Shuffling} In this strategy, we aim to randomly shuffle the order of the tokens in the input sequences. Since the bag-of-words nature in the transformer architecture, the position encoding is the only factor about the sequential information. Thus, similar to \citet{lee2020slm}, we implement this strategy by passing the shuffled position ids to the embedding layer while keeping the order of the token ids unchanged.

\textbf{Cutoff} \citet{shen2020simple} proposes a simple and efficient data augmentation strategy called cutoff. They randomly erase some tokens (for token cutoff), feature dimensions (for feature cutoff), or token spans (for span cutoff) in the $L \times d$ feature matrix. In our experiments, we only use token cutoff and feature cutoff and apply them to the token embeddings for view generation.

\textbf{Dropout} Dropout is a widely used regularization method that avoids overfitting. However, in our experiments, we also show its effectiveness as an augmentation strategy for contrastive learning. For this setting, we randomly drop elements in the token embedding layer by a specific probability and set their values to zero. Note that this strategy is different from Cutoff since each element is considered individually.

\begin{table*}[t]
\small
\centering
\resizebox{.9\textwidth}{!}{
\begin{tabular}{lcccccccc}
\toprule
& \textbf{STS12} & \textbf{STS13} & \textbf{STS14} & \textbf{STS15} & \textbf{STS16} & \textbf{STSb} & \textbf{SICK-R} & \textbf{Total} \\ \midrule
Number of train samples   & 0              & 0              & 0              & 0              & 0              & 5749          & 4500            & -              \\
Number of valid samples   & 0              & 0              & 0              & 0              & 0              & 1500          & 500             & -              \\
Number of test samples    & 3108           & 1500           & 3750           & 3000           & 1186           & 1379          & 4927            & -              \\
Number of Unlabeled Texts & 6216           & 3000           & 7500           & 17000          & 18366          & 17256         & 19854           & 89192          \\ \bottomrule
\end{tabular}}
\caption{The statistics of STS datasets.}
\label{tab:dataset_stat}
\vspace{-0.4cm}
\end{table*}

\subsection{Incorporating Supervision Signals}
\label{sec:incorporating_supervision_signals}

Besides unsupervised transfer, our approach can also be incorporated with supervised learning.
We take the NLI supervision as an example. It is a sentence pair classification task, where the model are trained to distinguish the relation between two sentences among \textit{contradiction}, \textit{entailment} and \textit{neutral}. The classification objective can be expressed as following:
\begin{equation}
    \setlength{\abovedisplayskip}{0.10cm}
    \setlength{\belowdisplayskip}{0.10cm}
    \begin{split}
    f = \text{Concat}(r_1, r_2, |r_1 - r_2|) \\
    \mathcal{L}_\text{ce} = \text{CrossEntropy}(Wf + b, y)
    \end{split}
\end{equation}
, where $r_1$ and $r_2$ denote two sentence representations.

We propose three ways for incorporating additional supervised signals:
\begin{itemize}
    \item \textbf{Joint training (joint)} We jointly train the model with the supervised and unsupervised objectives $\mathcal{L}_\text{joint} = \mathcal{L}_\text{ce} + \alpha \mathcal{L}_\text{con}$ on NLI dataset. $\alpha$ is a hyper-parameter to balance two objectives.
    \item \textbf{Supervised training then unsupervised transfer (sup-unsup)} We first train the model with $\mathcal{L}_\text{ce}$ on NLI dataset, then use $\mathcal{L}_\text{con}$ to fine-tune it on the target dataset.
    \item \textbf{Joint training then unsupervised transfer (joint-unsup)} We first train the model with the $\mathcal{L}_\text{joint}$ on NLI dataset, then use $\mathcal{L}_\text{con}$ to fine-tune it on the target dataset.
\end{itemize}

\section{Experiments}

To verify the effectiveness of our proposed approach, we conduct experiments on Semantic Textual Similarity (STS) tasks under the unsupervised and supervised settings.

\subsection{Setups}

\textbf{Dataset} Following previous works\cite{reimers2019sentencebert, li-etal-2020-sentence, zhang2020unsupervised}, we evaluate our approach on multiple STS datasets, including STS tasks 2012 - 2016 (STS12 - STS16) \cite{agirre2012semeval, agirre2013sem, agirre2014semeval, agirre2015semeval, agirre2016semeval}, STS benchmark (STSb) \cite{cer2017semeval} and SICK-Relatedness (SICK-R) \cite{marelli2014sick}. Each sample in these datasets contains a pair of sentences as well as a gold score between 0 and 5 to indicate their semantic similarity. For our unsupervised experiments, we mix the unlabeled texts from these datasets to fine-tune our model. We obtain all 7 datasets through the SentEval toolkit \cite{conneau2018senteval}. The statistics is shown in Table \ref{tab:dataset_stat}.

For supervised experiments, we use the combination of SNLI (570k samples) \cite{bowman2015large} and MNLI (430k samples) \cite{williams2018broad} to train our model. In the \textit{joint training} setting, the NLI texts are also used for contrastive objectives.

\textbf{Baselines} To show our effectiveness on unsupervised sentence representation transfer, we mainly select BERT-flow \cite{li-etal-2020-sentence} for comparison, since it shares the same setting as our approach. For unsupervised comparison, we use the average of GloVe embeddings, the BERT-derived native embeddings, CLEAR \cite{wu2020clear} (trained on BookCorpus and English Wikipedia corpus), IS-BERT \cite{zhang2020unsupervised} (trained on unlabeled texts from NLI datasets), BERT-CT \cite{carlsson2021semantic} (trained on English Wikipedia corpus). For comparison with supervised methods, we select InferSent \cite{conneau2017supervised}, Universal Sentence Encoder \cite{cer2018universal}, SBERT \cite{reimers2019sentencebert} and BERT-CT \cite{carlsson2021semantic} as baselines. They are all trained with NLI supervision.

\begin{table*}
\centering
\resizebox{.9\textwidth}{!}{
\begin{tabular}{lcccccccc}
\toprule
Method                      & STS12 & STS13 & STS14 & STS15 & STS16 & STSb  & SICK-R & Avg.  \\ \midrule
\multicolumn{9}{c}{\textit{Unsupervised baselines}}                              \\
Avg. GloVe embeddings$^\dagger$      & 55.14 & 70.66 & 59.73 & 68.25 & 63.66 & 58.02 & 53.76  & 61.32 \\
BERT$_{\text{base}}$$^\ddagger$                  & 35.20 & 59.53 & 49.37 & 63.39 & 62.73 & 48.18 & 58.60  & 53.86 \\
BERT$_{\text{large}}$$^\ddagger$                 & 33.06 & 57.64 & 47.95 & 55.83 & 62.42 & 49.66 & 53.87  & 51.49 \\
CLEAR$_{\text{base}}$$^\dagger$                 & 49.0 & 48.9 & 57.4 & 63.6 & 65.6 & 75.6 & 72.5  & 61.8 \\
IS-BERT$_{\text{base}}$-NLI$^\dagger$                  & 56.77 & 69.24 & 61.21 & 75.23 & 70.16 & 69.21 & 64.25  & 66.58 \\
BERT$_{\text{base}}$-CT$^\dagger$               & 66.86 & 70.91 & 72.37 & 78.55 & 77.78 & - & - & - \\
BERT$_{\text{large}}$-CT$^\dagger$               & 69.50 & 75.97 & 74.22 & 78.83 & 78.92 & - & - & - \\ \midrule
\multicolumn{9}{c}{\textit{Using STS unlabeled texts}}                        \\
BERT$_{\text{base}}$-flow$^\dagger$             & 63.48 & 72.14 & 68.42 & 73.77 & 75.37 & 70.72 & 63.11  & 69.57 \\
BERT$_{\text{large}}$-flow$^\dagger$            & 65.20 & 73.39 & 69.42 & 74.92 & \textbf{77.63} & 72.26 & 62.50  & 70.76 \\
% DeCLUTR$_{\text{base}}$$^\dagger$               & 64.21 & 70.40 & 69.99 & 77.51 & 75.35 & \textbf{78.84} & \textbf{79.39} & 73.67 \\
ConSERT$_{\text{base}}$$^\ddagger$               & 64.64 & 78.49 & 69.07 & 79.72 & 75.95 & 73.97 & 67.31  & 72.74 \\
ConSERT$_{\text{large}}$$^\ddagger$              & \textbf{70.69} & \textbf{82.96} & \textbf{74.13} & \textbf{82.78} & 76.66 & \textbf{77.53} & \textbf{70.37}  & \textbf{76.45} \\ \bottomrule
\end{tabular}}
\caption{The performance comparison of ConSERT with other methods in an \textit{unsupervised} setting. We report the spearman correlation $\rho \times 100$ on 7 STS datasets. Methods with $^\dagger$ indicate that we directly report the scores from the corresponding paper, while methods with $^\ddagger$ indicate our implementation.}
\label{tab:unsup_results}
\vspace{-0.4cm}
\end{table*}

\textbf{Evaluation} When evaluating the trained model, we first obtain the representation of sentences by averaging the token embeddings at the last two layers\footnote{As shown in \citet{li-etal-2020-sentence}, averaging the last two layers of BERT achieves slightly better results than averaging the last one layer.}, then we report the spearman correlation between the cosine similarity scores of sentence representations and the human-annotated gold scores. When calculating spearman correlation, we merge all sentences together (even if some STS datasets have multiple splits) and calculate spearman correlation for only once\footnote{Note that such evaluation procedure is different from SentEval toolkit, which calculates spearman correlation for each split and reports the mean or weighted mean scores.}.

\textbf{Implementation Details} Our implementation is based on the Sentence-BERT\footnote{\url{https://github.com/UKPLab/sentence-transformers}} \cite{reimers2019sentencebert}. We use both the BERT-base and BERT-large for our experiments. The max sequence length is set to 64 and we remove the default dropout layer in BERT architecture considering the \textit{cutoff} and \textit{dropout} data augmentation strategies used in our framework. The ratio of token cutoff and feature cutoff is set to 0.15 and 0.2 respectively, as suggested in \citet{shen2020simple}. The ratio of dropout is set to 0.2. The temperature $\tau$ of NT-Xent loss is set to 0.1, and the $\alpha$ is set to 0.15 for the joint training setting. We adopt Adam optimizer and set the learning rate to 5e-7. We use a linear learning rate warm-up over 10\% of the training steps. The batch size is set to 96 in most of our experiments. We use the dev set of STSb to tune the hyperparameters (including the augmentation strategies) and evaluate the model every 200 steps during training. The best checkpoint on the dev set of STSb is saved for test. We further discuss the influence of the batch size and the temperature in the subsequent section.

\subsection{Unsupervised Results}

For unsupervised evaluation, we load the pre-trained BERT to initialize the BERT encoder in our framework. Then we randomly mix the unlabeled texts from 7 STS datasets and use them to fine-tune our model.

The results are shown in Table \ref{tab:unsup_results}. We can observe that both BERT-flow and ConSERT can improve the representation space and outperform the GloVe and BERT baselines with unlabeled texts from target datasets. However, ConSERT$_\text{large}$ achieves the best performance among 6 STS datasets, significantly outperforming BERT$_\text{large}$-flow with an 8\% relative performance gain on average (from 70.76 to 76.45). Moreover, it is worth noting that ConSERT$_\text{large}$ even outperforms several supervised baselines (see Figure \ref{tab:sup_results}) like InferSent (65.01) and Universal Sentence Encoder (71.72), and keeps comparable to the strong supervised method SBERT$_\text{large}$-NLI (76.55). For the BERT$_\text{base}$ architecture, our approach ConSERT$_\text{base}$ also outperforms BERT$_\text{base}$-flow with an improvement of 3.17 (from 69.57 to 72.74).

\begin{table*}[t]
\centering
\resizebox{.9\textwidth}{!}{
\begin{tabular}{lcccccccc}
\toprule
Method                & STS12 & STS13 & STS14 & STS15 & STS16 & STSb  & SICK-R & Avg.  \\ \midrule
\multicolumn{9}{c}{\textit{Using NLI supervision}}                                              \\
InferSent - GloVe$^\dagger$          & 52.86 & 66.75 & 62.15 & 72.77 & 66.87 & 68.03 & 65.65  & 65.01 \\
Universal Sentence Encoder$^\dagger$ & 64.49 & 67.80 & 64.61 & 76.83 & 73.18 & 74.92 & 76.69  & 71.22 \\
SBERT$_{\text{base}}$-NLI$^\dagger$        & 70.97 & 76.53 & 73.19 & 79.09 & 74.30 & 77.03 & 72.91  & 74.89 \\
SBERT$_{\text{large}}$-NLI$^\dagger$       & 72.27 & 78.46 & 74.90 & 80.99 & 76.25 & 79.23 & 73.75  & 76.55 \\
SBERT$_{\text{base}}$-NLI (re-impl.)$^\ddagger$        & 69.89 & 75.77 & 72.36 & 78.51 & 73.67 & 76.75 & 72.76  & 74.24 \\
SBERT$_{\text{large}}$-NLI (re-impl.)$^\ddagger$       & 72.69 & 78.77 & 75.13 & 80.95 & 76.89 & 79.53 & 73.25  & 76.74 \\
BERT$_{\text{base}}$-CT$^\dagger$        & 68.80 & 74.58 & 76.62 & 79.72 & 77.14 & - & -  & - \\
BERT$_{\text{large}}$-CT$^\dagger$        & 69.80 & 75.45 & 76.47 & 81.34 & 78.11 & - & -  & - \\
ConSERT$_{\text{base}}$ \textit{joint}$^\ddagger$  & 70.53 & 79.96 & 74.85 & 81.45 & 76.72 & 78.82 & 77.53  & 77.12 \\
ConSERT$_{\text{large}}$ \textit{joint}$^\ddagger$ & \textbf{73.26} & \textbf{82.36} & \textbf{77.73} & \textbf{83.84} & \textbf{78.75} & \textbf{81.54} & \textbf{78.64}  & \textbf{79.44} \\ \midrule
\multicolumn{9}{c}{\textit{Using NLI supervision and STS unlabeled texts}}                      \\
BERT$_{\text{base}}$-flow$^\dagger$        & 68.95 & 78.48 & 77.62 & 81.95 & 78.94 & 81.03 & 74.97  & 77.42 \\
BERT$_{\text{large}}$-flow$^\dagger$       & 70.19 & 80.27 & 78.85 & 82.97 & \textbf{80.57} & 81.18 & 74.52  & 78.36 \\
ConSERT$_{\text{base}}$ \textit{sup-unsup}$^\ddagger$   & 73.51 & 84.86 & 77.44 & 83.11 & 77.98 & 81.80 & 74.29  & 79.00 \\
ConSERT$_{\text{large}}$ \textit{sup-unsup}$^\ddagger$ & 75.26 & \textbf{86.01} & 79.00 & 83.88 & 79.45 & 82.95 & 76.54  & 80.44 \\
ConSERT$_{\text{base}}$ \textit{joint-unsup}$^\ddagger$   & 74.07 & 83.93 & 77.05 & 83.66 & 78.76 & 81.36 & 76.77  & 79.37 \\
ConSERT$_{\text{large}}$ \textit{joint-unsup}$^\ddagger$ & \textbf{77.47} & 85.45 & \textbf{79.41} & \textbf{85.59} & 80.39 & \textbf{83.42} & \textbf{77.26}  & \textbf{81.28} \\ \bottomrule
\end{tabular}}
\caption{The performance comparison of ConSERT with other methods in a \textit{supervised} setting. We report the spearman correlation $\rho \times 100$ on 7 STS datasets. Methods with $^\dagger$ indicate that we directly report the scores from the corresponding paper, while methods with $^\ddagger$ indicate our implementation.}
\label{tab:sup_results}
\vspace{-0.4cm}
\end{table*}

\subsection{Supervised Results}

For supervised evaluation, we consider the three settings described in Section \ref{sec:incorporating_supervision_signals}. Note that in the \textit{joint} setting, only NLI texts are used for contrastive learning, making it comparable to SBERT-NLI. We use the model trained under the \textit{joint} setting as the initial checkpoint in the \textit{joint-unsup} setting. We also re-implement the SBERT-NLI baselines and use them as the initial checkpoint in the \textit{sup-unsup} setting.

The results are illustrated in Table \ref{tab:sup_results}. For the models trained with NLI supervision, we find that ConSERT \textit{joint} consistently performs better than SBERT, revealing the effectiveness of our proposed contrastive objective as well as the data augmentation strategies. On average, ConSERT$_\text{base}$ \textit{joint} achieves a performance gain of 2.88 over the re-implemented SBERT$_\text{base}$-NLI, and ConSERT$_\text{large}$ \textit{joint} achieves a performance gain of 2.70.

When further performing representation transfer with STS unlabeled texts, our approach achieves even better performance. On average, ConSERT$_\text{large}$ \textit{joint-unsup} outperforms the initial checkpoint ConSERT$_\text{large}$ \textit{joint} with 1.84 performance gain, and outperforms the previous state-of-the-art BERT$_\text{large}$-flow with 2.92 performance gain. The results demonstrate that even for the models trained under supervision, there is still a huge potential of unsupervised representation transfer for improvement.

\section{Qualitative Analysis}

\subsection{Analysis of BERT Embedding Space}
\label{sec:analysis_of_bert_embedding_space}

\begin{figure}[t]
    \centering
    \resizebox{.48\textwidth}{!}{
    \includegraphics{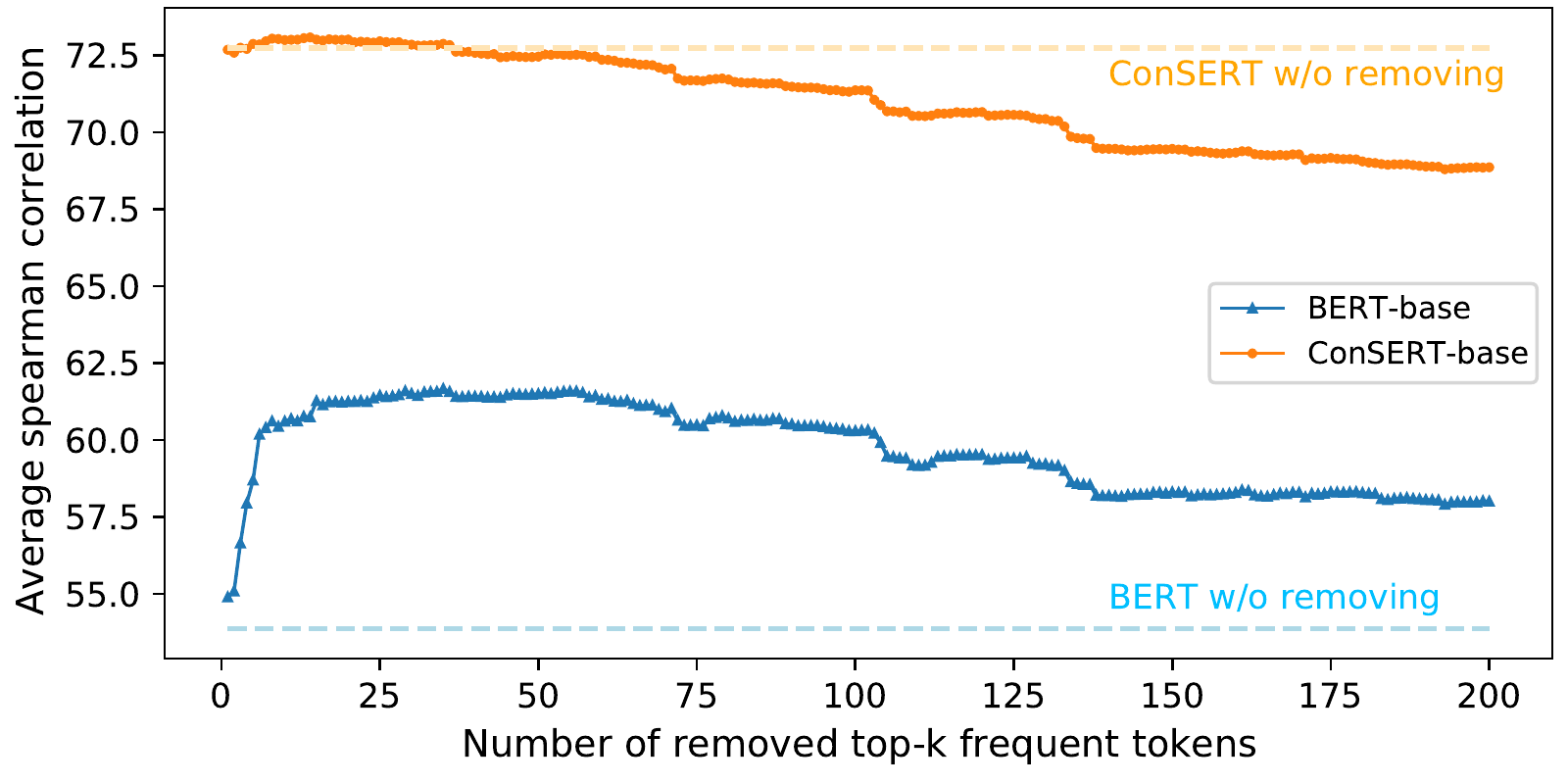}}
    \caption{The average spearman correlation on STS tasks w.r.t. the number of removed top-k frequent tokens. Note that we also considered the [CLS] and [SEP] tokens and they are the 2 most frequent tokens. The frequency of each token is calculated through the test split of the STS Benchmark dataset.}
    \label{fig:analysis_embeddings}
    \vspace{-0.4cm}
\end{figure}

To prove the hypothesis that the collapse issue is mainly due to the anisotropic space that is sensitive to the token frequency, we conduct experiments that mask the embeddings of several most frequent tokens when applying average pooling to calculate the sentence representations. The relation between the number of removed top-k frequent tokens and the average spearman correlation is shown in Figure \ref{fig:analysis_embeddings}.

We can observe that when removing a few top frequent tokens, the performance of BERT improves sharply on STS tasks. When removing 34 most frequent tokens, the best performance is achieved (61.66), and there is an improvement of 7.8 from the original performance (53.86). For ConSERT, we find that removing a few most frequent tokens only results in a small improvement of less than 0.3. The results show that our approach reshapes the BERT's original embedding space, reducing the influence of common tokens on sentence representations.

\subsection{Effect of Data Augmentation Strategy}

\begin{figure}[t]
    \centering
    \resizebox{.48\textwidth}{!}{
    \includegraphics{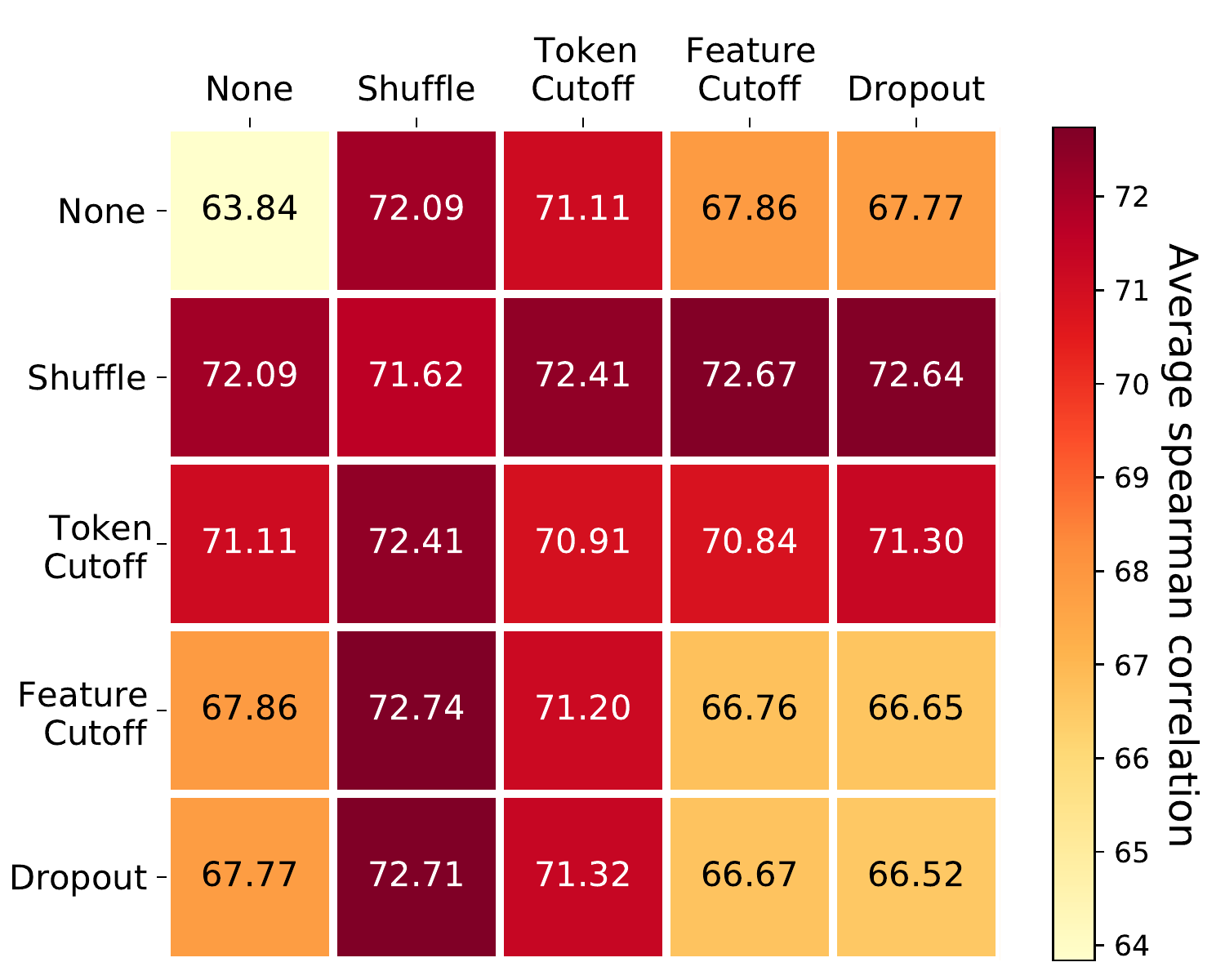}}
    \caption{The performance visualization with different combinations of data augmentation strategies. The row indicates the 1st data augmentation strategy while the column indicates the 2nd data augmentation strategy. }
    \label{fig:analysis_data_aug}
    \vspace{-0.3cm}
\end{figure}

In this section, we study the effect of data augmentation strategies for contrastive learning. We consider 5 options for each transformation, including None (i.e. doing nothing), Shuffle, Token Cutoff, Feature Cutoff, and Dropout, resulting in 5$\times$5 combinations. Note that the Adversarial Attack strategy is not considered here, since it needs additional supervision to generate adversarial samples. All these experiments follow the unsupervised setting and use the BERT$_\text{base}$ architecture.

The results can be found in Figure \ref{fig:analysis_data_aug}. We can make the following observations. First, Shuffle and Token Cutoff are the two most effective strategies (where Shuffle is slightly better than Token Cutoff), significantly outperforming Feature Cutoff and Dropout. This is probably because Shuffle and Token Cutoff are more related to the downstream STS tasks since they are directly operated on the token level and change the structure of the sentence to produce hard examples.

Secondly, Feature Cutoff and Dropout also improve performance by roughly 4 points when compared with the None-None baseline. Moreover, we find they work well as a complementary strategy. Combining with another strategy like Shuffle may further improve the performance. When combined Shuffle with Feature Cutoff, we achieve the best result. We argue that Feature Cutoff and Dropout are useful in modeling the invariance of the internal noise for the sentence encoder, and thus improve the model's robustness.

Finally, we also observe that even without any data augmentation (the None-None combination), our contrastive framework can improve BERT's performance on STS tasks (from 53.86 to 63.84). This None-None combination has no effect on maximizing agreement between views since the representations of augmented views are exactly the same. On the contrary, it tunes the representation space by pushing each representation away from others. We believe that the improvement is mainly due to the collapse phenomenon of BERT's native representation space. To some extent, it also explains why our method works.

\subsection{Performance under Few-shot Settings}

\begin{figure}
    \centering
    \resizebox{.48\textwidth}{!}{
    \includegraphics{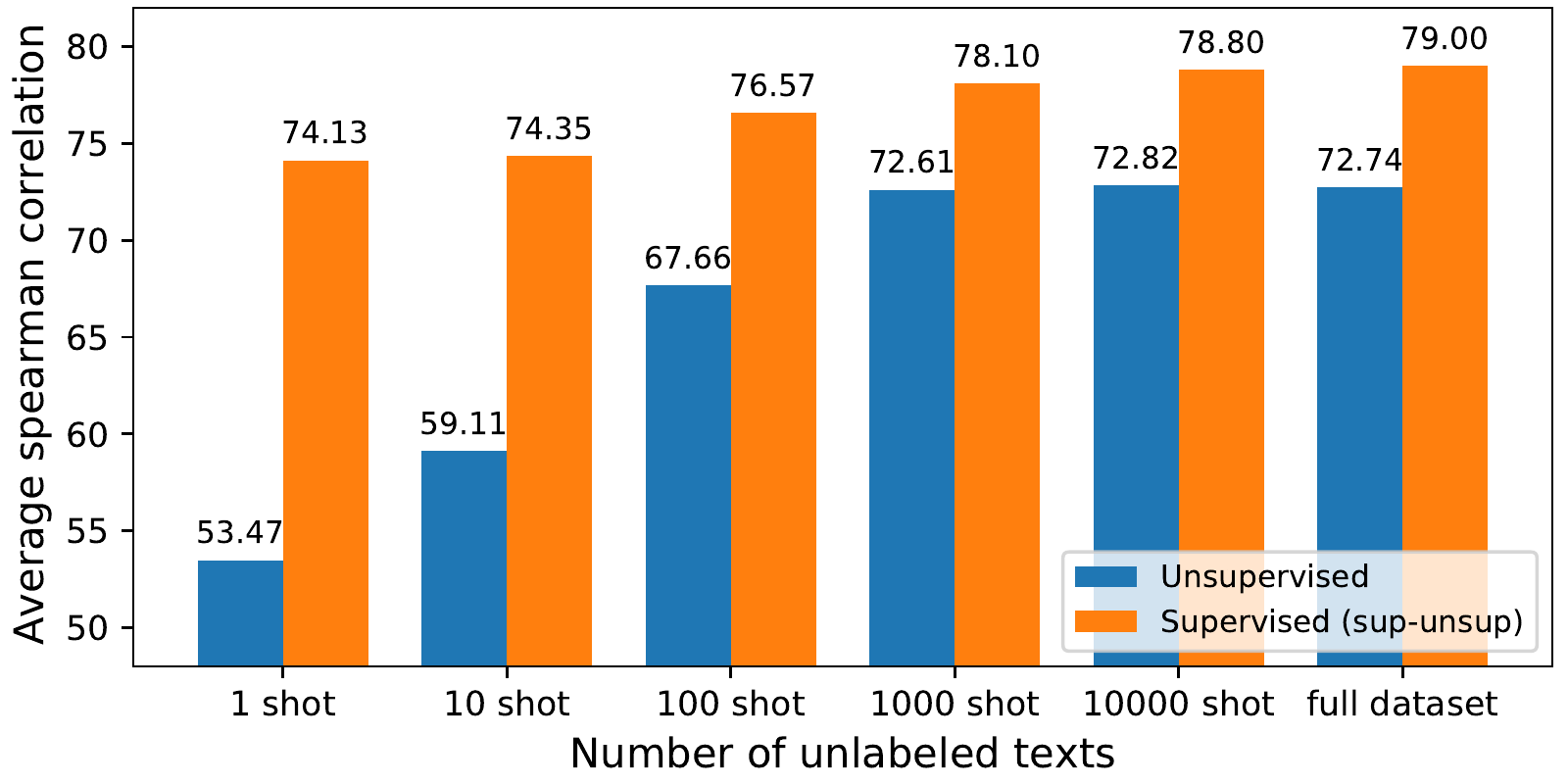}}
    \caption{The few-shot experiments under the unsupervised and supervised settings. We report the average spearman correlation on STS datasets with 1, 10, 100, 1,000, and 10,000 unlabeled texts available, respectively. The full dataset indicates all 89192 unlabeled texts from 7 STS datasets.}
    \label{fig:analysis_fewshot}
    \vspace{-0.3cm}
\end{figure}

To validate the reliability and the robustness of ConSERT under the data scarcity scenarios, we conduct the few-shot experiments. We limit the number of unlabeled texts to 1, 10, 100, 1000, and 10000 respectively, and compare their performance with the full dataset.

Figure \ref{fig:analysis_fewshot} presents the results. For both the unsupervised and the supervised settings, our approach can make a huge improvement over the baseline with only 100 samples available. When the training samples increase to 1000, our approach can basically achieve comparable results with the models trained on the full dataset. The results reveal the robustness and effectiveness of our approach under the data scarcity scenarios, which is common in reality. With only a small amount of unlabeled texts drawn from the target data distribution, our approach can also tune the representation space and benefit the downstream tasks.

\subsection{Influence of Temperature}

\begin{figure}[t]
    \centering
    \resizebox{.48\textwidth}{!}{
    \includegraphics{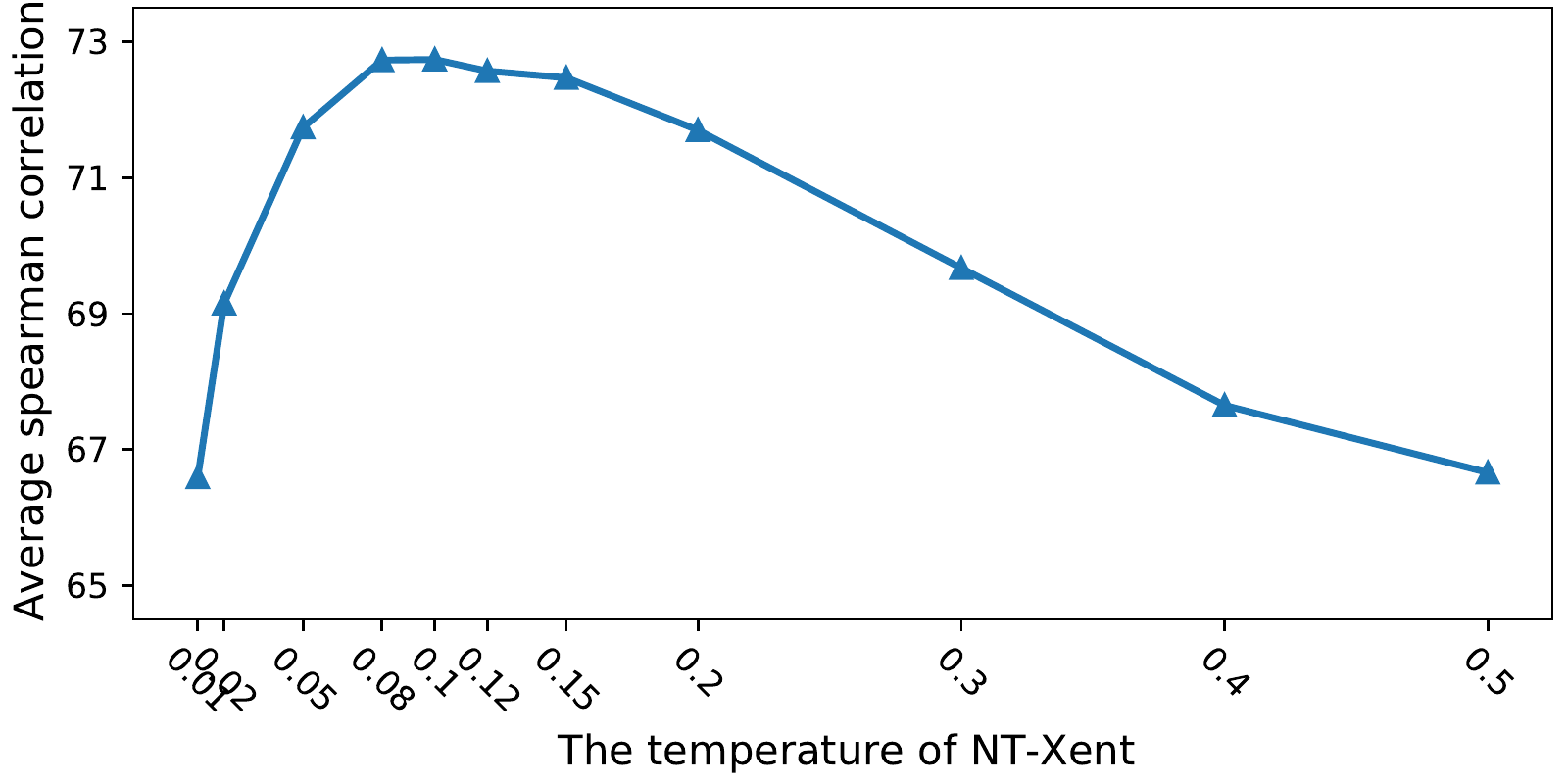}}
    \caption{The influence of different temperatures in NT-Xent. The best performance is achieved when the temperature is set to 0.1.}
    \label{fig:analysis_temperature}
    \vspace{-0.1cm}
\end{figure}

The temperature $\tau$ in NT-Xent loss (Equation \ref{equation:cl_loss}) is used to control the smoothness of the distribution normalized by softmax operation and thus influences the gradients when backpropagation. A large temperature smooths the distribution while a small temperature sharpens the distribution. In our experiments, we explore the influence of temperature and present the result in Figure \ref{fig:analysis_temperature}.

As shown in the figure, we find the performance is extremely sensitive to the temperature. Either too small or too large temperature will make our model perform badly. And the optimal temperature is obtained within a small range (from about 0.08 to 0.12). This phenomenon again demonstrates the collapse issue of BERT embeddings, as most sentences are close to each other, a large temperature may make this task too hard to learn. We select 0.1 as the temperature in most of our experiments.

\subsection{Influence of Batch Size}

In some previous works of contrastive learning, it is reported that a large batch size benefits the final performance and accelerates the convergence of the model since it provides more in-batch negative samples for contrastive learning \cite{chen2020simple}. Those in-batch negative samples improve the training efficiency. We also analyze the influence of the batch size for unsupervised sentence representation transfer.

The results are illustrated in Table \ref{tab:analysis_batch_size}. We show both the spearman correlation and the corresponding training steps. We find that a larger batch size does achieve better performance. However, the improvement is not so significant. Meanwhile, a larger batch size does speed up the training process, but it also needs more GPU memories at the same time.

\begin{table}
\centering
\resizebox{.48\textwidth}{!}{
\begin{tabular}{|c|c|c|c|c|c|}
\hline
\textbf{Batch Size}    & \textbf{16}    & \textbf{48}    & \textbf{96}    & \textbf{192}   & \textbf{288}   \\ \hline
\textbf{Avg. Spearman} & 72.63 & 72.60 & 72.74 & 72.86 & 72.98 \\ \hline
\textbf{Number of Steps} & 6175 & 2459 & 1530 & 930 & 620 \\ \hline
\end{tabular}}
\caption{The average spearman correlation as well as the training steps of our unsupervised approach with different batch sizes. }
\label{tab:analysis_batch_size}
\vspace{-0.3cm}
\end{table}

% \subsection{Analysis on Representation Space}

% \section{Limitation}

\section{Conclusion}

In this paper, we propose ConSERT, a self-supervised contrastive learning framework for transferring sentence representations to downstream tasks. The framework does not need extra structure and is easy to implement for any encoder. We demonstrate the effectiveness of our framework on various STS datasets, both our unsupervised and supervised methods achieve new state-of-the-art performance. Furthermore, few-shot experiments suggest that our framework is robust in the data scarcity scenarios. We also compare multiple combinations of data augmentation strategies and provide fine-grained analysis for interpreting how our approach works. We hope our work will provide a new perspective for future researches on sentence representation transfer.

\section*{Acknowledgements}

We thank Keqing He, Hongzhi Zhang and all anonymous reviewers for their helpful comments and suggestions. This work was partially supported by National Key R\&D Program of China No. 2019YFF0303300 and Subject II  No. 2019YFF0303302, DOCOMO Beijing Communications Laboratories Co., Ltd, MoE-CMCC ``Artifical Intelligence" Project No. MCM20190701.

\section*{Broader Impact}

Sentence representation learning is a basic task in natural language processing and benefits many downstream tasks. This work proposes a contrastive learning based framework to solve the collapse issue of BERT and transfer BERT sentence representations to target data distribution. Our approach not only provides a new perspective about BERT's representation space, but is also useful in practical applications, especially for data scarcity scenarios. When applying our approach, the user should collect a few unlabeled texts from target data distribution and use our framework to fine-tune BERT encoder in a self-supervised manner. Since our approach is self-supervised, no bias will be introduced from human annotations. Moreover, our data augmentation strategies also have little probability to introduce extra biases since they are all based on random sampling. However, it is still possible to introduce data biases from the unlabeled texts. Therefore, users should pay special attention to ensure that the training data is ethical, unbiased, and closely related to downstream tasks.

\bibliographystyle{acl_natbib}
\bibliography{anthology,acl2021}

\end{document}